\documentclass[10pt,twocolumn,letterpaper]{article}

\usepackage{cvpr}              

\usepackage{graphicx}
\usepackage{amsmath}
\usepackage{amssymb}
\usepackage{xcolor}
\usepackage{booktabs}
\usepackage{multirow}
\usepackage{enumitem}
\usepackage{array}
\usepackage[pagebackref,breaklinks,colorlinks]{hyperref}
\usepackage[toc,page]{appendix}

\usepackage[capitalize]{cleveref}
\crefname{section}{Sec.}{Secs.}
\Crefname{section}{Section}{Sections}
\Crefname{table}{Table}{Tables}
\crefname{table}{Tab.}{Tabs.}

\newcolumntype{x}[1]{>{\centering\arraybackslash\hspace{0pt}}p{#1}}

\definecolor{mybrown}{RGB}{197, 90, 17}
\definecolor{mygreen}{RGB}{125, 160, 66}
\definecolor{mypurple}{RGB}{112, 48, 160}
\definecolor{myyellow}{RGB}{255, 217, 102}
\definecolor{mygray}{RGB}{215, 215, 215}
\definecolor{mydarkgray}{RGB}{127, 127, 127}
\definecolor{myblue}{RGB}{126, 148, 184}
\newcommand\Loss{\mathcal{L}}

\makeatletter
\newcommand{\printfnsymbol}[1]{%
  \textsuperscript{\@fnsymbol{#1}}%
}

\def\cvprPaperID{8619} 
\def\confName{CVPR}
\def\confYear{2022}

\begin{document}
\title{ContIG: Self-supervised Multimodal \underline{Cont}rastive Learning for\\ Medical \underline{I}maging with \underline{G}enetics}

\author{Aiham Taleb$^1$\thanks{Equal contribution}
\qquad 
Matthias Kirchler$^{1,2}$\printfnsymbol{1}
\qquad
Remo Monti$^1$
\qquad
Christoph Lippert$^{1,3}$ \\ 
\small
$^1 $ Hasso Plattner Institute for Digital Engineering, University of Potsdam, Germany  \\
\small
$^2 $ TU Kaiserslautern, Germany  \\
\small
$^3 $ Hasso Plattner Institute for Digital Health at the Icahn School of Medicine at Mount Sinai, NYC, USA  \\
{\tt\small \{firstname.lastname\}@hpi.de} \\
}

\maketitle

\begin{abstract}
    High annotation costs are a substantial bottleneck in applying modern deep learning architectures to clinically relevant medical use cases, substantiating the need for novel algorithms to learn from unlabeled data.
    In this work, we propose ContIG, a self-supervised method that can learn from large datasets of unlabeled medical images and genetic data.
    Our approach aligns images and several genetic modalities in the feature space using a contrastive loss.
    We design our method to integrate multiple modalities of each individual person in the same model end-to-end, even when the available modalities vary across individuals.
    Our procedure outperforms state-of-the-art self-supervised methods on all evaluated downstream benchmark tasks.
    We also adapt gradient-based explainability algorithms to better understand the learned cross-modal associations between the images and genetic modalities.
    Finally, we perform genome-wide association studies on the features learned by our models, uncovering interesting relationships between images and genetic data.
\end{abstract}

\section{Introduction} \label{sec:intro}
\begin{figure}[ht]
  \centering
  \includegraphics[width=\linewidth]{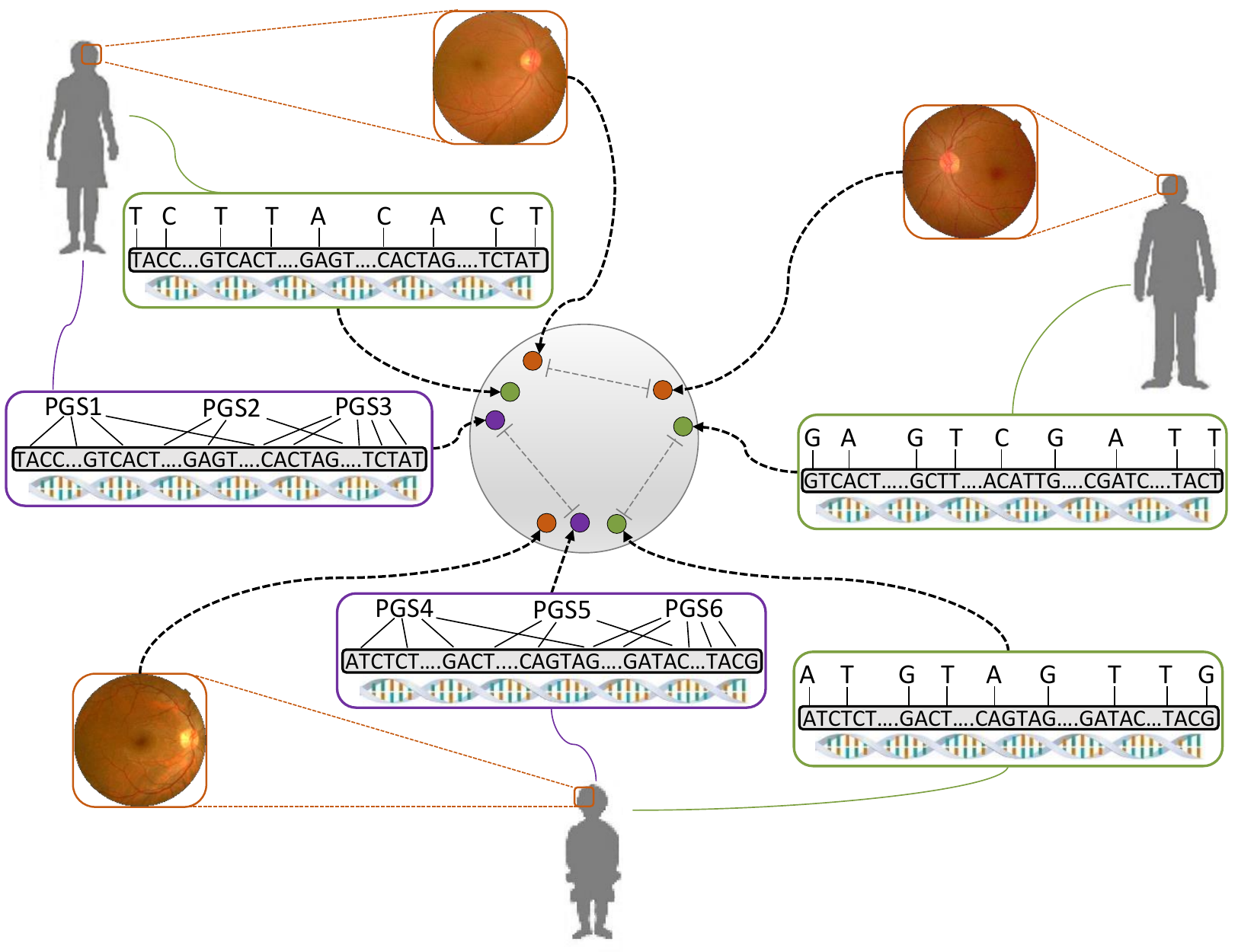}
  \caption{Overview of our contrastive learning method from imaging and genomic data. It learns representations by bringing the modalities of each individual closer in the embedding space, and apart from different individuals'. In this example, the modalities are retinal fundus images (in \textcolor{mybrown}{brown}), SNP data (in \textcolor{mygreen}{green}), and polygenic risk scores (PGS) (in \textcolor{mypurple}{purple}). Our method handles missing modalities (\eg missing PGS for the person in the upper right).}
  \label{fig:teaser}
\end{figure}

Medical imaging plays a vital role in patient healthcare. It aids in disease prevention, early detection, diagnosis, and treatment. However, efforts to employ machine learning algorithms to support in clinical settings are often hampered by the high costs of required expert annotations~\cite{annotate}. 
At the same time, large-scale biobank studies have recently started to aggregate unprecedented scales of multimodal data on human health. For example, the UK Biobank (UKB)~\cite{ukbio} contains data on $500,000$ individuals, including a wide range of imaging modalities such as retinal fundus images and cardiac, abdominal, and brain MRI. Similar studies are currently underway in other countries, such as the Nationale Kohorte (NaKo)~\cite{nako}, BioMe~\cite{biome}, FinnGen~\cite{FinnGen}, Estonia Biobank~\cite{est_biobank}, and others. 
While some of these studies also include phenotypic descriptions, \eg a person's medical history, such data tend to be both highly incomplete and biased due to clinical practices and assessment methods~\cite{o2005measuring}, making learning from them challenging and error-prone.
On the other hand, genetic data is increasingly abundant. While chip-based genotyping technology has enabled the study of common genetic variation at scale~\cite{verlouw2021comparison}, the exponentially decreasing costs of genomic sequencing is driving progress for rare genetic variation~\cite{park2016trends}. Due to these advances, the UKB and other biobanks often contain a rich array of genetic and genomic measurements. Genetic data is generally less susceptible to bias factors, and most diseases have at least a partially genetic cause, with some genetic disorders being exclusively attributed to genetic mutations~\cite{witte2014contribution}.
Similarly, most other traits -- not directly related to diseases --, \eg height and human personality, are also strongly influenced by genetics~\cite{lippert2017identification,zwir2020uncovering}.
Complementary imaging-genetics datasets are increasingly also available in other application settings, \eg plant breeding~\cite{yang2020crop}.

Unlabelled medical images carry valuable information about organ structures, and an organism's genome is the blueprint for biological functions in the individual's body. Clearly, integrating these distinct yet complementary data modalities can help create a more holistic picture of physical and disease traits. 
Integrating these data types, however, is non-trivial and challenging.
The human genome consists of three billion base pairs, yet most genetic differences between individuals have little effect. This leads to challenges both in terms of computational aspects, and in terms of statistical efficiency.
Unfortunately, it is not clear a priori which parts of the genome are relevant and which are not. Typically, genome-wide association studies (GWAS)~\cite{gwas1,gwas2} use statistical inference techniques to discover relationships between genetic variations and particular physical or disease traits.
To date, thousands of scientific works have found more than $300,000$ genetic-phenotype associations~\cite{gwas_cat}.
However, even now a large portion of known or presumed heritability of traits is not yet accounted for by the individual genome-trait associations, a phenomenon known as ``missing heritability''~\cite{manolio2009finding}.
Better methods to find -- and explain -- the relationships between genetic and imaging modalities may help close this gap.

Therefore, the growing number of biobanks of \textit{unlabeled} multimodal (\ie imaging-genetics) data, calls for solutions that can: (i) learn semantic data representations without requiring expensive expert annotations, (ii) integrate these data modalities end-to-end in an efficient manner, and (iii) explain discovered cross-modal correspondences (associations). Self-supervised (unsupervised) representation learning offers a viable solution when unlabeled data is abundant and labels are scarce. These methods witnessed a surge of interest after proving successful in several application domains~\cite{survey_self_supervised}. The representations learned by these methods facilitate data-efficient fine-tuning on supervised downstream tasks, reducing significantly the burden of manual annotation. Furthermore, such methods allow for integrating multiple data modalities as distinct views, which can lead to considerable performance gains.
Despite the recent advancements in self-supervised methods, \eg contrastive learning, only little work has been done to adopt these methods in the medical domain. In fact, we are not aware of any prior work that leverages self-supervised representation learning on combined imaging and genetic modalities. 
We believe self-supervised learning has the potential to address the above challenges inherent to the medical domain. 

\noindent\textbf{Contributions.} 
\textbf{(i)} We propose a self-supervised method, called ContIG, that can learn from multimodal datasets of \textit{unlabeled} medical images and genetic data. ContIG aligns these modalities in the representation space using a contrastive loss, which enables learning semantic representations in the same model end-to-end. Our approach handles the case of multiple genetic modalities, in conjunction with images, even when the available modalities vary across individuals.
\textbf{(ii)} We adapt gradient-based explainability algorithms to better understand the learned cross-modal correspondences (associations) between the images and genetic modalities. Our method discovers interesting associations, and we confirm their relevance by cross-referencing biomedical literature.

Our work presents a framework on how to exploit inexpensive self-supervised solutions on large corpora (\eg Biobanks) of (medical) images and genetic data.

\section{Related Work} \label{sec:related_work}
\textbf{Self-supervised learning with pretext tasks.} These methods learn an embedding (representation) space by deriving a proxy (pretext) task from the data itself, requiring no human labels. The learned embeddings will also be useful for real-world downstream tasks, afterwards. A large body of works relied on such proxy tasks~\cite{w2v,context_prediction,jig,color,deep_cluster,rotations}. A comprehensive review of similar works is provided in~\cite{survey_self_supervised}. The limitation of such methods is the need to design handcrafted proxy tasks to learn representations.

\textbf{Contrastive learning} approaches~\cite{CPC1,CPC2,simclr,byol,simsiam,moco,dim,pirl,wu2018unsupervised,barlow,swav,mocov2,nnclr} circumvent the above challenge by maximizing mutual information between related signals in contrast to others, by employing Noise Contrastive Estimation~\cite{NCE}. Contrastive methods advanced the results of unsupervised learning on ImageNet~\cite{imagenet_cvpr09}. However, unlike our method, these methods process uni-modal images only.  

\textbf{Multimodal learning.} Learning from multimodal data poses several inherent challenges, such as: multimodal fusion, alignment, and representation~\cite{multimodal_survey,multimodal_dl}. Prior works, some of which are self-supervised, learn from a variety of modalities, such as: image and text (vision and language)~\cite{videobert,cbt,vilbert,lxmert,visualbert,image_caption1,image_caption2,cross_modal_scenes}, image and audio~\cite{audio_image1,audio_image2,audio_image3,audio_image4,audio_image5,soundnet}, audio and text~\cite{audio_text1,audio_text2}, and multi-view (multimodal) images~\cite{cmc,cross_learn,rgb_optical1,rgb_optical2}. More recent works employed contrastive learning for multimodal inputs (image and text captions)~\cite{contrastive_sound,contrastive_text1,contrastive_text2,contrastive_text3,contrastive_text4,contrastive_text_audio}. We follow this line of work, and we extend contrastive pretraining to novel modalities, \ie images and genetics, for the first time.

\textbf{Self-supervision on medical images.}
Early works of self-supervision in the medical context~\cite{depth,surgery,register,body,disc,Yan2018DeepLG,cardiac_self_supervised,endoscopic_videos,Cytoarchitectonic_segmentation} made assumptions about input data, limiting their generalization to other target tasks. Then, many works proposed employing proxy tasks for self-supervision from medical scans~\cite{orientation_prediction_tajbakhsh,spitzer_3d_distance,ssl_models_genesis,ultrasound_video,taleb_multimodal_puzzles,rubik,image_context_medical,ultrasound_video,image_context_medical2}. A review of similar works is in~\cite{review_annotation_efficient}. Recently, contrastive learning~\cite{chaitanya2020contrastive,taleb_3D_paper,contrastive_registration1,contrastive_registration2} has been applied to medical scans, where it also showed promising results. Our work, as opposed to these works, utilizes multiple modalities (images and one or more genetic modalities) to improve the learned representations by capturing imaging-genetic relationships.

\textbf{Deep learning from both genetics and images.} In addition to its successful applications to medical imaging~\cite{dl_medical_imaging}, deep learning also found success in applications on genomics~\cite{dl_in_genetics1,dl_in_genetics2,dl_in_genetics3,dl_in_genetics4,dl_in_genetics5}.
There is a growing number of recent works that utilize deep neural networks to jointly learn from both modalities, such as ~\cite{joint_ad1,joint_ad2,multimodal_ad, ash2021joint,gundersen2020end,badea2020identifying,transferGWAS,chang2018deep,fujinami2021prediction,dai2021multi}.
However, these methods are all either highly application specific or fully supervised.
Notably, we are not aware of any prior work leveraging the self-supervised framework (with contrastive loss functions) to improve representation learning from both imaging and genetic data.

\section{Method} \label{sec:method}
In~\cref{sec:gen_primer} we first review some biomedical foundations and motivate the genetic modalities chosen in this work. 
Then, we describe our contrastive method in~\cref{sec:con_method}, and different modality aggregation types. Finally, we detail the explanation methods for genetic features in~\cref{sec:gen_explain}.

\subsection{Modalities of Genetic Data} \label{sec:gen_primer}
The basic building blocks of DNA, which encodes the biological functions needed for the development of an organism, are called nucleotides. A long sequence of the four nucleotides Adenine (A), Thymine (T), cytosine (C), and Guanine (G) make up the genome - the "recipe" needed to build an organism~\cite{dna_seq}.
A relatively small fraction of the genome codes for proteins, while the remaining parts have regulatory or structural functions. 
However, over generations, genetic mutations occur, for example substituting one nucleotide for another, \eg A to C. Some of these genetic changes can alter physical traits (\eg eye color), or cause disease (\eg Alzheimer's).
"Genotyping" is the process of measuring these genetic changes~\cite{snp_genotyping}.
The most frequently measured type of changes are single-nucleotide-polymorphisms (SNPs), where a single pair of nucleotides is altered at a specific position in the genome.

There are three billion base pairs in the human genome, but typically only a small fraction of them is measured, due to cost and technological restraints. 
Even if large parts of the sequence are available, as is the case for whole-genome sequencing studies, working with the raw data is not feasible, both in terms of \emph{statistical efficiency} -- most of those base pairs carry no causal signal and only add noise to the estimation process -- and in terms of \emph{computational efficiency}.
For these reasons, most studies record only a small subset of all nucleotides, usually on the order of several hundred thousand to several million SNPs.
Furthermore, human traits of interest are constructed by a spectrum of different genetic architectures.
At the same time, due to evolutionary dynamics, some SNPs exhibit their possible variations frequently in a population (``common'' variants), while other SNPs are identical for the overwhelming majority of the population with only few individuals having mutations (``rare'' variants) -- a form of class imbalance.
Therefore, in this work we consider \emph{three} different ways to encode the genetic modalities that emphasize different aspects of human physiology.

\emph{Complex traits} are traits that are influenced by a large number of causal factors, including relatively common genetic variations.
One example is height, which is determined to a large degree by many SNPs all across the human genome~\cite{yang2010common}.
Many common diseases and impairments are complex traits, which makes them especially relevant to human health applications~\cite{frazer2009human}.
To best encode genetic architectures associated with complex traits, we utilize \textbf{polygenic risk scores (PGS)}~\cite{pgs}.
PGS aggregate many, mostly common, SNPs into a single score that reflects a person's inherited susceptibility to a specific disease~\cite{pgs_nature}.
The individual SNPs are weighted based on their strength of association with the disease.
By using many different PGS for different traits and diseases we can get a multi-faceted view of an individual's complex trait predisposition.

Recent advances in DNA sequencing have also enabled assessing the contribution of \emph{rare genetic variants} to heritable traits~\cite{rare}.
Rare variants occur at low frequencies (\eg MAF\footnote{minor allele frequency} $<1\%$ or MAF $\ll 1\%$) in a population.
Large genetic effects often negatively affect an individual's health and are strongly selected against by evolution.
Hence, in contrast to common variants, many rare variants have a large effect size and predispose for genetic diseases.
Rare variants are usually not included in PGS, and due to their low frequencies they pose a challenge for robust statistical models.
In this work, we use \textbf{burden scores}~\cite{burden_test}, which aggregate several rare variants within a localized genetic region. 

Finally, we also employ a uniformly sampled cross section of the whole genome, by including every $k$-th SNP that has been genotyped in the respective study. 
These \textbf{raw SNPs} are mostly common variants (due to the biological sampling procedure) and give a broad representation of an individual's genetic composition.
This representation likely carries population structure such as ancestry~\cite{lippert2011fast}, but also tags highly diverse functional information.

These three genetic modalities -- polygenic risk scores, burden scores, and raw SNPs -- capture complementary aspects and together paint a broad description of an individual's genetic predisposition.
We employ them both individually and jointly as contrastive views to medical images.

\begin{figure*}[ht]
  \centering
  \includegraphics[width=\linewidth]{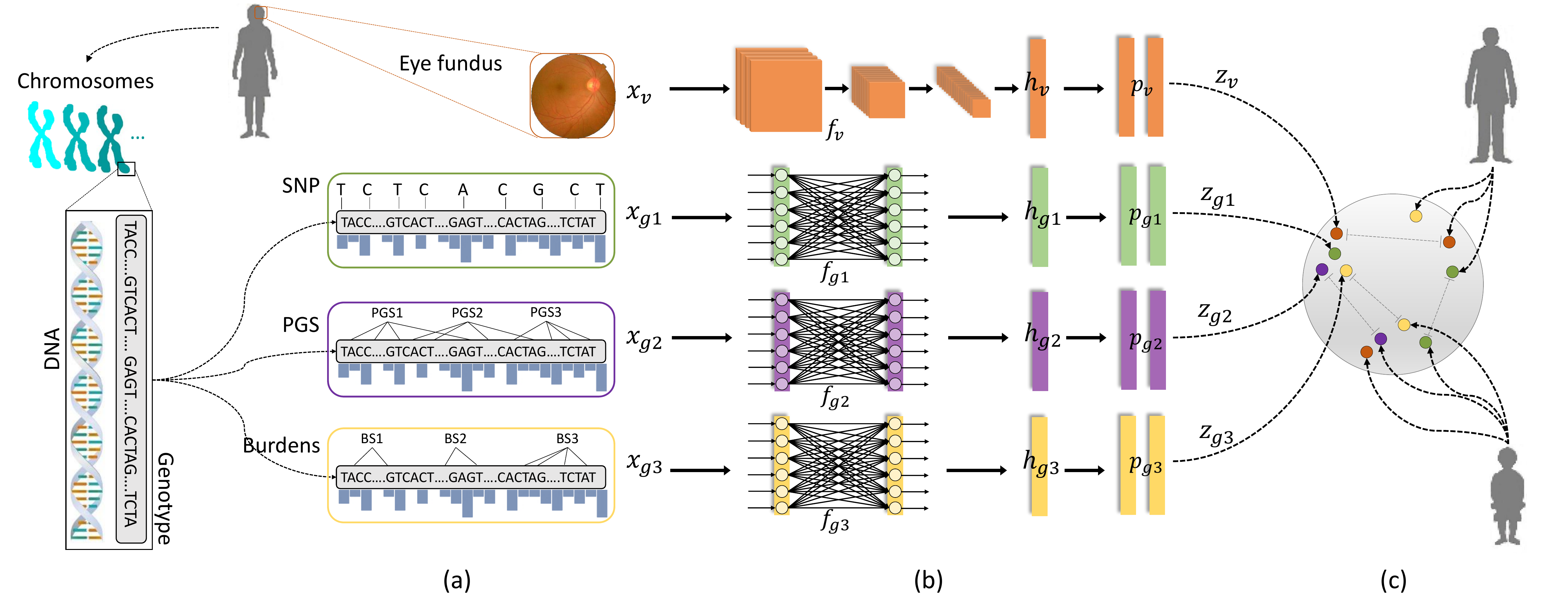}
  \caption{Schematic illustration for the steps of our proposed method. \textbf{(a)} Assuming one imaging modality (retinal fundus shown in \textcolor{mybrown}{brown}), and three genetic modalities (Single-nucleotide polymorphisms (SNP) in \textcolor{mygreen}{green}, polygenic risk scores (PGS) in \textcolor{mypurple}{purple}, burden scores in \textcolor{myyellow}{yellow}). Note that different genetic modalities exhibit different variant frequencies (denoted by the histogram in \textcolor{myblue}{blue}): SNP and PGS use common variants (high frequency), while burdens use rare variants (low frequency).  \textbf{(b)} We extract features from each modality with deep neural networks, \ie Convolutional Networks for images and Fully Connected (MLP) networks from genomic data. We use a projection head (MLP) for each modality, which produces equally-sized modality embeddings $z_v, z_{g1}, z_{g2}, z_{g3}$. \textbf{(c)} We use these embeddings in the contrastive loss computation. The embeddings of each individual are encouraged to come closer in the feature space (depicted by the \textcolor{mygray}{gray} circle), and farther from other individuals'. The dotted \textcolor{mydarkgray}{gray} lines demonstrate the contrasting mechanism between modalities.  
  }
  \label{fig:method}
\end{figure*}

\subsection{Contrastive Learning from Images \& Genetics} \label{sec:con_method} 
We assume a dataset of $N$ multimodal samples, one for each individual person. Each sample consists of a medical image paired with multiple genetic modalities. Here, we denote each image by $x^{i}_{v}$, and the corresponding genetic modalities by $x^{i}_{gm}$, where $i \in \{1,..,N\}$ is the individual and $m \in \{1,..,M\}$ is the genetic modality.
We group images and genetic modalities in batches of size $b>1$ by the individual modalities: $v = \{x_v^{i_1}, \ldots, x_v^{i_b}\}$ and $g_m := \{x_{gm}^{i_1}, \ldots, x_{gm}^{i_b}\}$. The number of available genetic modalities may vary across individuals. 

Our method, illustrated in~\cref{fig:method}, processes these input modalities with a set of neural network encoders, one per modality. We denote the image encoding by $h^i_{v} = f_{v}(x^{i}_{v})$, and the genetics encodings as $h^i_{gm} = f_{gm}(x^{i}_{gm})$, with $M$ distinct genetics encoders. The resulting $d$-dimensional vector representations $h^i_{v}, h^i_{gm} \in \mathbb{R}^d$ are then processed with projection heads $z^i_v=p_v(h^i_{v}), z^i_{gm}=p_{gm}(h^i_{gm})$, respectively, where $z_v, z_{gm} \in \mathbb{R}^d$. Following~\cite{simclr}, each projection head is a non-linear MLP with one hidden-layer.  

\paragraph{Contrastive Loss with Two Modalities.} 
We first define the contrastive loss assuming $N$ pairs of an image and one genetic modality $(x^i_v, x^i_g)$, with their respective representations $(z^i_v, z^i_g)$. Then, for the image sample in the $i^{th}$ pair, we consider the genetic sample $x^i_g$ as the positive (true) sample among the negative genetic samples of other individuals $x^k_g$ in the same batch.
Similarly, the image $x^i_v$ is the positive sample of $x^i_g$, amongst the negative image samples $x^k_v$. Therefore, the contrastive loss is the sum of these two parts: i) image-to-genetics $L(v,g)$ (fix the image and contrast genetic samples), and ii) genetics-to-image $L(g,v)$ (fix the genetics and contrast images). Formally, in each step of the training we select a random batch of size $b > 1$ with indices $\{ i_1, \ldots, i_b \}$ and use the batch-wise loss function: 
\begin{equation}
\begin{split}
    L(v, g) &= 
    - \sum_{j=1}^{b} \log \frac{
            \exp(\cos(z^{i_j}_v, z^{i_j}_g)/\tau)
        }{
            \sum^b_{k=1, k\neq j} \exp(\cos(z^{i_j}_v,z^{i_k}_g)/\tau)
        } 
    \\
   \Loss_{cont}(v,g) &= \lambda L(v, g) + (1-\lambda) L(g, v),
\end{split}
  \label{eq:loss-one-gen}
\end{equation}
where $\tau$ is a temperature parameter, $\cos$ is the cosine similarity, and $\lambda \in [0,1]$ is a loss weighting hyperparameter. 

\paragraph{Generalizing to Multiple Genetic Modalities.} 
We generalize here the above contrastive loss formulation to the case when there are multiple available genetic modalities, corresponding to the same image sample. Since we aim to improve the learned visual representations mainly, the image modality is used at the center of this training scheme (we deem alternative contrasting schemes a future work). 
In other words, we contrast the image with each one of the $M$ genetic modalities. Therefore, the generalized multimodal contrastive loss becomes:
\begin{equation}
   \Loss(v, g_1, \ldots, g_M) = \sum^M_{m=1} \Loss_{cont}(v, g_m)
  \label{eq:loss-all-gen}
\end{equation}
This formulation ensures the learned visual representations capture useful information from all available genetic modalities. However, this assumes that every individual has all the genetic modalities, which is not normally the case. Hence, we define two aggregation schemes to handle the missing genetic modalities: i) the "inner" aggregation scheme, which uses only those individuals for which \textit{all} the modalities exist, and ii) the "outer" aggregation scheme, which covers all the individuals, even those with \textit{missing} genetic modalities.
In particular, for each $\Loss_{cont}(v, g_m)$ in~\cref{eq:loss-all-gen}, the ``outer'' aggregation only includes individuals with non-missing data for this specific modality. 
The "outer" scheme can make better use of all available data. 
Both schemes allow for training on combinations of existing modalities.

\subsection{Genetic Features Explanation} \label{sec:gen_explain}
For a given multimodal tuple $x := (x_v, x_{g1}, \ldots, x_{gM})$ of image and genetic representations, we perform feature explanations to understand the contribution of each genetic feature $g_{m, j}$ for the model output.
Standard deep learning explainability approaches are not directly applicable in this setting, as they require a simple one-to-one relation from input to output, while the contrastive loss \cref{eq:loss-all-gen} is computed over batches. 
Instead, we utilize a fixed reference batch of $b \ge 1$ individuals with images $v_r$ and genetic modalities $g_{m,r}$ ($m=1, \ldots, M$) and define the explainer function
\begin{equation*}
	E(x) := \Loss( v_r \cup \{x_v\}, g_{1, r} \cup \{ x_{g1} \}, \ldots, g_{M, r} \cup \{ x_{gM} \} )
\end{equation*}
with $\Loss$ defined as in ~\cref{eq:loss-all-gen}, but $v_r, g_{1,r}, \ldots, g_{M,r}$ fixed.
We can then use standard feature attribution methods such as Integrated Gradients~\cite{integrated_gradients} or DeepLift~\cite{deep_lift} to explain the contribution of all elements in $x$ towards the full batch loss.
We can additionally also fix the input image $x_v$ to only consider the attribution of the genetic effects.
Note that the explanation will be sensitive to the choice of the reference batch; to minimize this effect, we choose $b$ to be relatively large ($b=1,000$ in our experiments).

In addition to these \emph{local} instance-specific attributions, we are especially interested in understanding the behavior of our models \emph{globally}.
For this, we aggregate many individual explanations, all using the same (independent) reference batch.
Feature importance both in negative and positive direction is important in our setting, and therefore we consider the mean absolute value for each feature dimension as a measure of global attribution. 

The setting with missing values can be handled analogously to the "outer" aggregation scheme in \cref{sec:con_method}, by just omitting the respective modalities. 

\section{Experimental Results} \label{sec:results}
In this section, we present the evaluation results of our method. First, we detail the datasets used for pretraining and evaluation (\cref{sec:datasets}). Then, we assess the quality of the learned representations (\cref{sec:downstream}), by: i) fine-tuning (\ie transfer learning) on four downstream tasks, and ii) performing a genome-wide association study (GWAS) on the model features. Finally, we present the genetic feature explanation results (\cref{sec:explain_results}), and we analyze the findings to check their relevance with medical literature resources.

\subsection{Datasets} \label{sec:datasets}
We pretrain our models (and the unsupervised baselines) on data obtained from the UK Biobank (UKB) dataset~\cite{ukbio}. This dataset contains multimodal data for almost 500k individuals, although imaging data is only available for a subset of those.
The UKB consists to an overwhelming majority of individuals of European descent; we restrict our dataset to those European descent individuals, as including more individuals would likely introduce very large confounding effects~\cite{lippert2011fast}.
For the purposes of pretraining, we utilize the retinal fundus images, which amount to $155,238$ imaging samples (left and right eyes). The genetic modalities we employ (see~\cref{sec:gen_primer}), amount to $155,238$ Raw-SNP samples (all individuals have Raw-SNPs), $145,206$ PGS samples, and $93,216$ burden scores. In terms of feature dimensions, for the raw-SNPs, we uniformly sample every $100^{th}$ SNP from $22$ Chromosomes (excluding the X and Y chromosomes), resulting in $7,854$ SNPs per sample. For PGS, we used 481 scores for a wide variety of different traits downloaded from the PGS Catalog~\cite{lambert2021polygenic}.
We created burden scores for $18,574$ protein-coding genes~\cite{monti2021identifying}. These binary scores indicate whether a participant has at least one potentially damaging rare (MAF $< 1\%$) variant within a given gene.
We holdout a test split ($20\%$) from the UKB dataset, and the remaining data are for training ($70\%$) and validation ($10\%$). Each person only appears in one split.

For the downstream tasks, we employ: 
i) APTOS 2019 Blindness Detection~\cite{APTOS} dataset for Diabetic Retinopathy detection in~\cref{sec:aptos}, which has $3,662$ retinal fundus training samples. 
ii) Retinal Fundus Multi-disease Image Dataset (RFMiD)~\cite{rfmid} for disease classification (\cref{sec:rfmid}), which has $3,200$ training images. iii) $102,219$ images from the UKB~\cite{ukbio} training split, but now we predict cardiovascular risk factors (\cref{sec:ukb-cov}). 
iv) Pathologic Myopia challenge dataset~\cite{palm} for Pathological Myopia Segmentation (\cref{sec:palm}), which has $400$ image samples with segmentation masks. More datasets details in the appendix.

\subsection{Transfer Learning Results} \label{sec:downstream} 
In this section, we evaluate the quality of representations by fine-tuning to downstream tasks. However, we find that a linear evaluation protocol~\cite{CPC1,simclr} (encoder weights are kept frozen) behaves similarly to fine-tuning, see appendix.

\paragraph{Models \& architectures.} Across the following experiments, we employ a Resnet50~\cite{resnet} architecture as the encoder for image data ($f_v$ in~\cref{fig:method}). For the genetic encoders ($f_{gm}$), we vary the number of fully connected layers: "None" hidden layers, one hidden layer "H1", and two hidden layers "H12". We also vary the combination of genetic modalities (detailed in \cref{sec:gen_primer}) used in pretraining, along with modality aggregation schemes (explained in \cref{sec:con_method}). 

\paragraph{Baselines.} We compare to the following baselines:
\begin{itemize}[noitemsep]
    \item Training from scratch (\textbf{Baseline}): we train the same model on each downstream task, but initialized from random weights. Comparing to this baseline provides insights about the benefits of pretraining.
    \item State-of-the-art contrastive methods: we compare to self-supervised (contrastive) methods from literature by training on the same data splits, and using the same experimental setup. Namely, we compare to models pretrained with \textbf{SimCLR}~\cite{simclr}, \textbf{BYOL}~\cite{byol}, \textbf{Barlow Twins}~\cite{barlow}, \textbf{SimSiam}~\cite{simsiam}, and \textbf{NNCLR}~\cite{nnclr}.
\end{itemize}

\begin{table*}
  \centering
  \begin{tabular}[t]{ l p{2cm} x{2cm} x{2cm} x{2cm} x{2cm} @{\hspace{-0.1cm}}c } \toprule
    \multicolumn{2}{c}{\multirow{2}{*}{Model \& Genetics Encoder}} & \multicolumn{1}{c}{APTOS} & \multicolumn{1}{c}{RFMiD} & \multicolumn{1}{c}{PALM} & \multicolumn{2}{c}{Cardio. Risk Pred.} \\
    \cline{3-7}
     & & QwKappa $\uparrow$ & ROC-AUC $\uparrow$ & Dice-Score $\uparrow$ & MSE $\downarrow$ & ROC-AUC $\uparrow$ \\
    \hline 
    Baseline                    & - & 80.47 & 91.64 & 77.25 & 3.440 & 56.29 \\
    \hline
    SimCLR~\cite{simclr}        & - & 81.83 & 91.88 & 70.41 & 3.451 & 59.38 \\
    SimSiam~\cite{simsiam}      & - & 75.44 & 91.28 & 72.26 & 3.442 & 57.37 \\
    BYOL~\cite{byol}            & - & 71.09 & 89.88 & 66.32 & 3.414 & 59.73 \\
    Barlow Twins~\cite{barlow}  & - & 72.28 & 92.03 & 70.53 & 3.430 & 59.05 \\
    NNCLR~\cite{nnclr}          & - & 77.93 & 91.89 & 72.06 & 3.426 & 61.95 \\
    \hline
    ContIG (Raw-SNP)    & H1 & 84.01 & 93.22 & 76.98 & 3.254 & 70.10 \\
    ContIG (PGS)        & H1 & \textbf{85.93} & \underline{93.31} & \textbf{78.47} & \underline{3.176} & \textbf{72.72}\\
    ContIG (Burden)     & H1 & 83.22 & 93.03 & 76.49 & \textbf{3.160} & \underline{72.37} \\
    ContIG (Inner RPB)  & H1 & 81.52 & 92.95 & \underline{77.34} & 3.202 & 70.80 \\
    ContIG (Outer RPB)  & H1 & \underline{84.22} & \textbf{93.62} & 76.97 & 3.187 & 71.80\\
    \bottomrule
  \end{tabular}
  \caption{Downstream evaluation results by fine-tuning on each task. \textbf{Bold} indicates the best result, \underline{underlined} is second best. RPB in our method stand for the genetic modalities used: Raw-SNPs, PGS-scores, and Burden-scores. $\uparrow$ means higher is better, and $\downarrow$ lower is better.}
  \label{tab:downstream}
\end{table*}

\subsubsection{Diabetic Retinopathy Detection (APTOS)} \label{sec:aptos}
Millions of people suffer from Diabetic Retinopathy, the leading cause of blindness among working aged adults. The APTOS dataset~\cite{APTOS} contains 2D fundus images, which were rated by a clinician on a severity scale of $0$ to $4$. These levels define a five-way classification task. We fine-tune the image encoder of our models and the baselines on this dataset, and then we evaluate on a fixed test split ($20\%$ of the data). The metric used in the task, as in the official Kaggle challenge, is the Quadratic Weighted Kappa (QwKappa~\cite{qwkappa}), which measures the agreement between two ratings. Its values vary from random (0) to complete agreement (1), and if there is less agreement than chance it may become negative. The evaluation results in~\cref{tab:downstream} support the effectiveness of our proposed contrastive method (ContIG). Our pretrained models outperform all baselines in this task, demonstrating the quality of its learned representations.

\subsubsection{Retinal Fundus Disease Classification (RFMiD)} \label{sec:rfmid}
The Retinal Fundus Multi-disease Image Dataset (RFMiD)~\cite{rfmid} also contains 2D fundus images, which are captured using three different cameras. It has 46 class labels, which represent disease conditions annotated through adjudicated consensus of two experts. 
Similarly, to evaluate on this task, we fine-tune the image encoders on this dataset, and we measure the performance on the test set. We should note that this task is solved as a multi-label classification task, since the patients may have multiple conditions at the same time. As an evaluation metric, we compute area under the ROC curve (ROC-AUC), and we use a micro averaging scheme~\cite{micro_roc_auc}. The results for this task in~\cref{tab:downstream} also demonstrate the gains in performance obtained by training with ContIG. Our models also outperform the self-supervised baselines in this task. 

\subsubsection{Pathological Myopia Segmentation (PALM)} \label{sec:palm}
Myopia has become a global burden of public health. Pathologic myopia causes irreversible visual impairment to patients, which can be detected by the changes it causes in the optic disc, including peripapillary atrophy, tilting, etc. The PALM dataset~\cite{palm} contains segmentation masks for these lesions, from which we evaluate on disc and atrophy segmentation tasks. Similar to the above downstream tasks, we fine-tune the image encoder on this dataset and evaluate on the test split. To predict segmentation masks, we add a u-net decoder~\cite{UNET} on top of the ResNet50 encoder. In terms of evaluation metrics, we use the dice score~\cite{dice_score}. The results of this task in~\cref{tab:downstream} demonstrate the quality of the learned representations by ContIG on semantic segmentation.

\subsubsection{Cardiovascular Risk Prediction} \label{sec:ukb-cov}
Previous work has shown that retinal fundus images can predict a range of risk factors for cardiovascular diseases~\cite{google_cardio_prediction}. Namely, retinal fundus images have been found to carry information about age, sex, smoking status, systolic and diastolic blood pressure (SBP, DBP), and body mass index (BMI). We predict these six risk factors using a subset of the UK Biobank~\cite{ukbio} dataset, by fine-tuning the image encoder on these values. As evaluation metrics, we use Mean Squared Error (MSE) for the numerical factors (age, BMI, SBP, DBP), and we use the ROC-AUC value for the categorical factors (sex and smoking status). As~\cref{tab:downstream} shows, models pretrained with ContIG outperform the baseline models in both classification and prediction tasks.

\subsubsection{Genome-wide Association Study Results} \label{sec:gwas} 
A GWAS is a statistical analysis that correlates individual genetic markers sampled along the full genome with a trait of interest, such as a specific disease.
GWASs usually require a low-dimensional, well-defined trait for association analysis; there is only little work yet on leveraging full medical imaging data in a GWAS setting~\cite{ash2021joint,transferGWAS}.
Here, we follow the \emph{transferGWAS}~\cite{transferGWAS} framework to evaluate the embeddings learned by ContIG.
In this framework, images are projected onto their latent space embeddings and then the dimensionality is further reduced with a Principal Component Analysis.
These low dimensional image representations can then be efficiently associated with SNPs using statistical association analysis tools such as PLINK~\cite{purcell2007plink,chang2015second}.
To compare different training methods, we count how many independent genetic regions each method finds; a more expressive image representation is expected to find more associated regions.
We defer the analysis details to the appendix. 

\cref{tab:gwas} shows the number of found independent regions for each pretraining method.
Genetic pretraining increases the statistical power of the genetic association study considerably.
Only BYOL~\cite{byol} achieves near-competitive results and all other self-supervised methods are outperformed by a large margin.  
We also looked up the found regions in the GWAS catalog \cite{gwas_cat} of published association results.
Many of the regions were already known to be associated with skin pigmentation.
This is not surprising, as the retina is known to be pigmented itself, which again is likely to be correlated with actual skin pigmentation.
Besides pigmentation, the GWAS catalog records associations with an array of cardiovascular traits (such as BMI, pulse pressure, large artery stroke, and blood biomarkers), as well as eye-specific associations (cataract and astigmatism).
Similar results were found by \cite{transferGWAS}, albeit with a larger sample size.

\begin{table}
  \centering
  \begin{tabular}[t]{ l c } \toprule
     Model		& Found Regions $\uparrow$ \\
    \hline 
    SimCLR~\cite{simclr}		    & 4 \\
    SimSiam~\cite{simsiam}		    & 2 \\
    BYOL~\cite{byol}                & 17 \\
    Barlow Twins~\cite{barlow}      & 8 \\
    NNCLR~\cite{nnclr}              & 3 \\
    \hline
    ContIG (Raw-SNP)               & 16 \\
    ContIG (PGS)                   & \underline{20} \\
    ContIG (Burden)                & 19 \\
    ContIG (Inner RPB)			       & \textbf{22} \\
    ContIG (Outer RPB)			       & 18 \\
    \bottomrule
  \end{tabular}
  \caption{GWAS results. Indicated is the number of independent regions associated with the image embeddings for each model}
  \label{tab:gwas}
\end{table}

\subsection{Genetic Feature Explanation Results} \label{sec:explain_results} 
\begin{figure}[t]
  \centering
  \includegraphics[width=\linewidth]{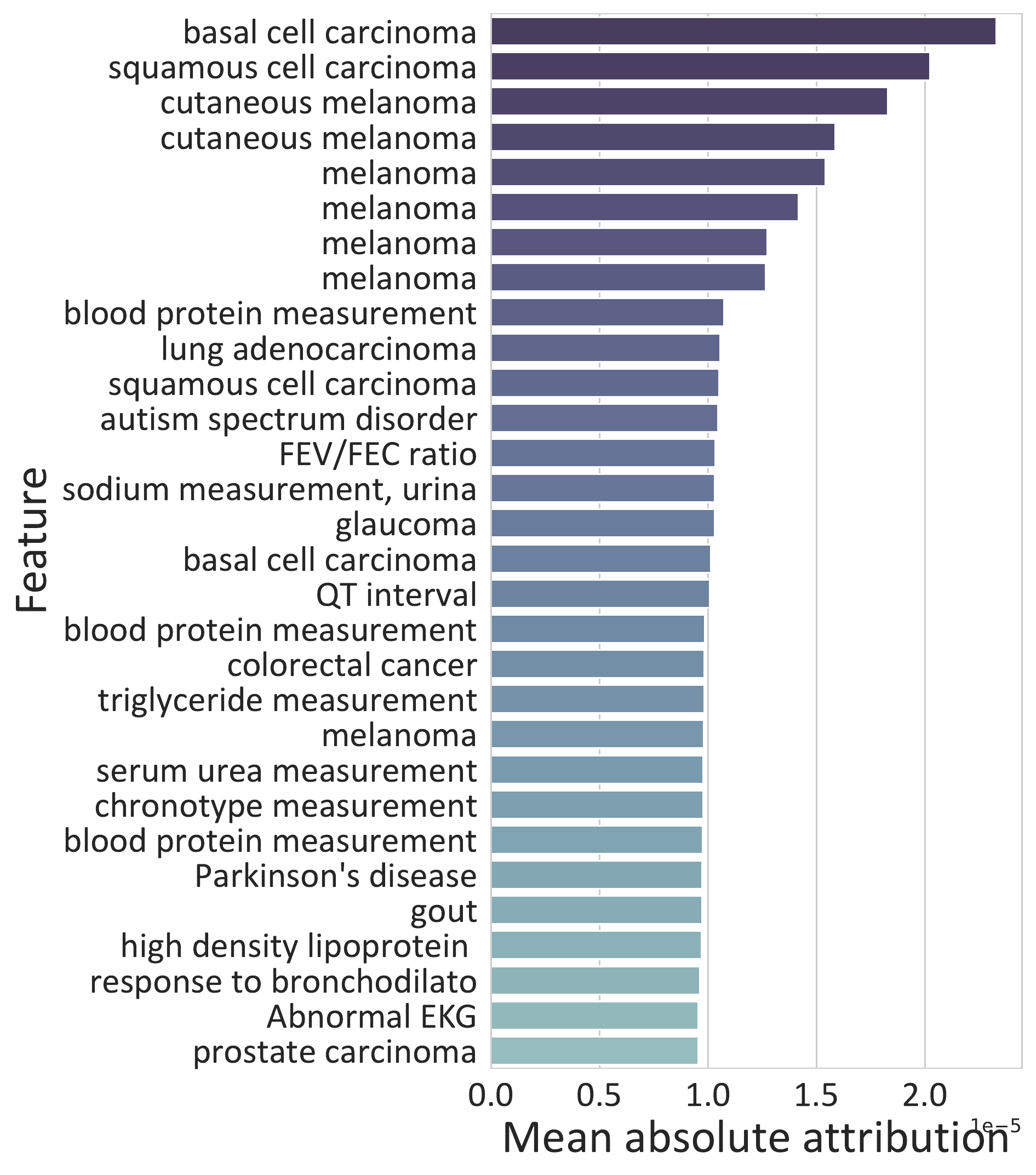}
  \caption{Global explanations for genetic features in ContIG (PGS only).
  Recorded is the mean absolute attribution per feature, aggregated over 1000 individuals, and the 30 PGS with highest associations are shown.
  Repeated traits (\eg Melanoma) are due to multiple different risk scores published in the PGS catalog.}
  \label{fig:global-exp}
\end{figure}

In this section, we investigate the representations learned by ContIG, using the explanation methods developed in \cref{sec:gen_explain}.
We first validate that our explanation approach can distinguish meaningful features from noise features, see appendix. 
Next, we analyze the models trained with a single genetic modality.
\cref{fig:global-exp} shows the 30 PGS with the strongest attributions, aggregated over 1,000 examples with a reference batch of size 1,000.
The most important features are different kinds of skin cancers (basal \& squamous cell carcinoma, cutaneous melanoma and melanoma).
This can be explained by the fact that the retina is pigmented and skin pigmentation is highly correlated to skin cancer.

Besides that, glaucoma, which is a disease of the optic nerve, is a highly relevant PGS, and many of the other traits are linked to cardiovascular functions (abnormal EKG, HDL cholesterol, blood protein measurements, QT interval), smoking status (lung adenocarcinoma, FEV/FEC ratio, response to bronchodilator) and liver and kidney function (triglyceride \& serum urea measurements). 
This is in line with previous studies which found strong signals with similar biomarkers in retinal fundus images~\cite{google_cardio_prediction}.
Interestingly, ContIG also finds correlations with neurological conditions such as Parkinson's disease and autism, which have previously been linked to retinal changes as well \cite{satue2014retinal,gialloreti2014reduction}.

Similarly, among the 15 strongest associations for raw SNPs, these SNPs were previously associated with cardiovascular traits (rs10807207, rs228416, rs1886785, rs10415889, rs3851381), pigmentation (rs228416), neurological and psychological conditions (rs1886785, rs1738895, rs6533374), and smoking status (rs6533374).

\begin{figure}[t]
  \centering
  \includegraphics[width=\linewidth]{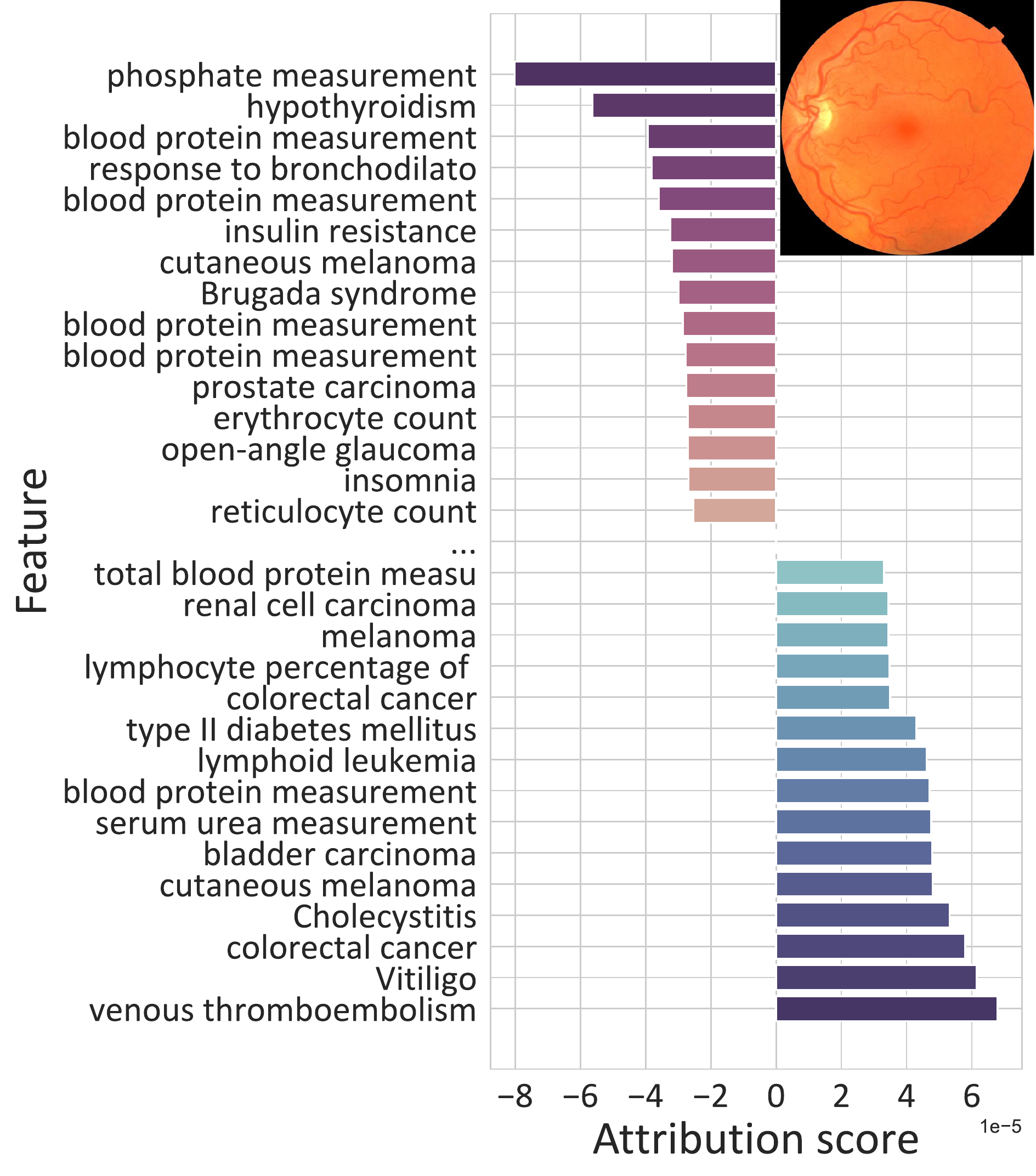}
  \caption{Local explanation attributions (signed) of genetic features for one image-PGS pair. Only the risk scores with highest values in either positive direction are shown. Retinal fundus image reproduced by kind permission of UK Biobank ©.}
  \label{fig:local-exp}
\end{figure}

In addition to the global attributions, \cref{fig:local-exp} shows the local attributions for one image/PGS pairing.
The retinal fundus image shows strong signs of vascular tortuosity, a known and important biomarker for cardiovascular conditions \cite{cheung2011retinal}.
Analogously, for this instance there is a large number of PGSs very strongly related to cardiovascular health (insulin resistance, many blood biomarkers, type II diabetes, Brugada syndrome, thromboembolism). 

These local and global explanations together provide further evidence that self-supervised pretraining with ContIG is able to learn semantically meaningful image representations without the need for manual annotations.
We provide additional explanatory results in the appendix.

\section{Discussion \& Limitations}  \label{sec:conclusion}
We presented ContIG, a self-supervised representation learning algorithm for imaging-genetics datasets.
Our evaluation results show that including genetic information in the pretraining process can considerably boost performance of image models in a variety of downstream tasks relevant for clinical practice and genetic research. 
We additionally conjecture that the self-supervised baseline methods' reliance on image augmentations alone may be disadvantageous in medical applications
due to the more uniform nature (\eg color distributions) of medical images compared to in natural images. 
We also leveraged interpretability methods to understand the relationship between imaging and genetic modalities in more detail and find interesting associations.

Naturally, there are a number of limitations for our proposed approach.
First, ContIG requires datasets that capture both imaging and genetics data, and is thus not applicable to pure-imaging datasets.
In recent years, however, an increasing number of imaging-genetics studies have started, and proprietary datasets of joint imaging and genetics data are available in some large-scale health systems. With the decreasing prices in both imaging and genotyping technology, this trend is likely to continue further.
A second limitation lies in the potentially limited application fields of our method. ContIG is not applicable to standard natural images, as there are no corresponding genetic features.
On the other hand, large-scale biobanks often include multiple imaging modalities, such as different MRI and histopathology images. Our method is also applicable to imaging-genetics applications in live-stock and plant breeding, and may also be useful in basic science studies.

Unfortunately, most large-scale imaging-genetics datasets to date are conducted in European and Northern American countries. 
Therefore, one limitation of the presented results is that the UKB mostly consists of populations with European ancestry, and may carry a biased representation.
We have shown that ContIG nevertheless improves downstream tasks in other populations, \eg in APTOS (collected in India), RFMiD (collected in India), and PALM (collected in China). We deem extending ContIG to other medical imaging datasets and genetic populations a future work.

{\small
\bibliographystyle{ieee_fullname}
\bibliography{egbib}
}

 \clearpage
\begin{appendices}
\section{Training \& Implementation Details}

\subsection{Imaging Preprocessing}

\subsubsection{Image Quality Control}
\label{sec:image-qc}
The UK Biobank contains a relatively large number of retinal fundus images with bad quality (\eg completely black or extremely overexposed).
To filter out extreme outliers, we performed two steps of quality control.
First, we only included images where a simple circle-detection algorithm \cite{illingworth1987adaptive} could find a circle.
In the second step, we filtered out the top and bottom $0.5\%$ brightest and darkest remaining images.

\subsubsection{Image transformations}
We cropped images to the circles detected in \cref{sec:image-qc} and rescaled to $448\times 448$ pixels.
During training, we randomly transform images by a rotation of up to $20^{\circ}$ and flip the image horizontally with a $50\%$ probability. We also follow the common practice of normalizing (standardizing) all the image intensities using the mean and standard deviation from ImageNet~\cite{imagenet_cvpr09}.

\subsection{Genetics Preprocessing} 
In all our experiments we used the genetic data provided by the UK Biobank.
The three different genetic modalities require different preprocessing steps, which we detail in this section.
\subsubsection{Raw SNPs}
\label{sec:raw-prep}
The raw SNPs are a cross section of all SNPs collected on microarray chips, collecting approximately 800k genetic variants in total across all chromosomes.
More information on data collection can be found at \url{https://biobank.ctsu.ox.ac.uk/crystal/label.cgi?id=263}.

The individual SNPs are coded in additive format, \ie 0 stands for no deviation from the reference genome, 1 means that one of the two chromosome copies has a deviation and the other not, and 2 means that both chromosome copies show a deviation from the reference genome.
We treated SNPs as continuous variables (opposed to, \eg separating them into three classes each) and imputed missing values by mode imputation.
Since 800k feature dimensions are challenging to handle, and SNPs are highly spatially correlated along the genome~\cite{reich2001linkage}, we only sampled every 100-th SNP from the full microarray.
We also only included SNPs on the 22 autosomal (=not sex-specific) chromosomes, as handling sex chromosomes requires special statistical care and leads to non-shared features between genetic males and females.
Together, this means we include 7,854 SNPs in our models.

\subsubsection{Polygenic Risk Scores}
\label{sec:pgs-prep}
For computing polygenic risk scores, we downloaded all PGS weight files included in the PGS Catalog~\cite{lambert2021polygenic} (\url{https://ftp.ebi.ac.uk/pub/databases/spot/pgs/}, last accessed October 11, 2021; at the time of writing, a large batch of new scores has been added to the PGS catalog), a collection of published PGS.
The PGS files provide weights for a linear model to compute risk scores from the raw genetic data.
To have a large intersection of available SNPs for our UKB population and the weights provided by the PGS catalog, instead of using the raw microarray data from \cref{sec:raw-prep}, we used \emph{imputed} data.
The imputed data uses prior knowledge about correlations between SNPs collected and not collected on the respective microarray (``linkage disequilibrium'', LD) to infer the missing features with high accuracy.
Imputed data was precomputed by the UKB, and more information can be found at \url{https://biobank.ctsu.ox.ac.uk/crystal/label.cgi?id=100319}.
Using the imputed data, we computed 481 polygenic scores for our cohort using the PLINK software \cite{purcell2007plink}, ignoring scores that gave errors or that only recorded genome positions in a different reference genome build.

For some traits, there are multiple distinct risk scores in the PGS catalog, as multiple independent studies have been performed on the same trait.
For example, the trait ``melanoma'' appears 9 times in our subset of selected PGS scores, while other traits, such as ``insomnia'' appear only once.
The scores contain partially overlapping genetic markers, and the number of SNPs used for each individual score vary from only 1 to several millions.

\subsubsection{Burden Scores}
We ran the Functional Annotation and Association Testing Pipeline 
\cite{monti2021identifying} to functionally annotate all the genetic variants present in the UK Biobank 200k exome sequencing release \cite{szustakowski2021advancing}.  Protein loss of function and missense variants that were predicted to be damaging were used to construct burden scores across all protein coding genes. We considered only rare variants with minor allele frequencies below {1\%}. Of these variants {41\%} were "singletons", i.e. only observed once in our sample. Specifically, each participant was assigned a binary vector of length 18,574 corresponding to the number of protein coding genes.  For every gene, the entry in this vector is 1 if the participant harbored at least one potentially damaging variant in that gene, or 0 if no potentially damaging variants were observed in that gene for that participant. This coding has been applied in rare-variant association studies in order to aggregate the effects of many rare variants within genes, where it can boost statistical power and reduce the burden of multiple testing\cite{burden_test, monti2021identifying}.

\subsection{Downstream Datasets Preprocessing}
\subsubsection{Diabetic Retinopathy detection (APTOS)}
In this task we use the APTOS 2019 Blindness Detection~\cite{APTOS} dataset, which has $3,662$ retinal fundus training samples. As explained in the main paper, the labels in this dataset have five levels of disease severity, defining five classes. However, these classes are not mutually exclusive, as a higher disease severity of \eg four is also of level three and below. Hence, we employ a multi-hot encoding scheme for the labels. For instance, class three is encoded as \texttt{[1,1,1,0,0]} and two as \texttt{[1,1,0,0,0]}, and so on. We split the dataset into three different splits of training (60\%), validation (20\%), and test (20\%). There is no overlap of patients across these splits.

\subsubsection{Retinal Fundus Disease Classification (RFMiD)}
For this task, we use the Retinal Fundus Multi-disease Image Dataset (RFMiD)~\cite{rfmid}, which has $3,200$ images. The overall number of disease classes is 45. However, we found that two classes ("HR" and "ODPM") have no positive cases, so we exclude these two classes and only work with the remaining 43 classes. As mentioned before, we convert these classes to multi-hot labels and solve the task as multilabel classification. We use this dataset's official splits for training, validation, and test. 

\subsubsection{Pathological Myopia Segmentation (PALM)}
We use the Pathologic Myopia challenge dataset~\cite{palm} for this task, which has 400 image samples with segmentation masks. As for segmentation labels, this dataset has three annotated areas: i) peripapillary atrophy (available for 311 cases), ii) optic disc (available for all cases), and iii) detachment (available for 12 cases only). Given that detachment is rarely available, we omit it from this task and only predict the atrophy and disc classes. We stratify the patients using the atrophy labels, to ensure equal representation of classes in train (60\% of dataset size) / val (20\%) / test (20\%) splits.

\subsubsection{Cardiovascular Risk Prediction (UKB)}
To predict the cardiovascular risk factors of (sex, age, BMI, SBP, DBP, smoking status) from retinal fundus scans, we use $102,219$ images from the UKB~\cite{ukbio}.
This corresponds to the training split ($70\%$ of UKB dataset size). We use the remaining scans for validation ($10\%$ of dataset size) and for the test split ($20\%$). Each person only appears in one split. The training for this task is performed using two models: i) one model to classify the categorical labels (sex to binary labels \{0,1\}, smoking status to binary labels too), ii) a second model to predict -- solved as a regression task -- the remaining continuous variables (age, BMI, SBP, and DBP). We use two models because the loss values of these two tasks have different scales. We preprocess the values of the continuous factors by standardization (removing the mean and scaling to unit variance). Finally, we impute the missing values of these factors by using the "mean" for continuous factors and "median" for discrete factors.

\subsection{Training Details}
We provide the training details for all pretraining (self-supervised) and downstream tasks in this section. 
\begin{itemize}[noitemsep]
    \item \textbf{Batch sizes}: we use a unified batch size of 64 across all pretraining and downstream tasks.
    \item \textbf{Optimizers}: we use Adam optimizer~\cite{Adam_supp} in all pretraining and downstream tasks. 
    \item \textbf{Schedulers}: during self-supervised pretraining (with ContIG and the baselines), we decay the learning rate with the cosine decay schedule without restarts~\cite{cosine_annealing}.
    \item \textbf{Learning rates}: we use an initial learning rate of $0.001$ across all tasks. However, we reduce the learning rate during training in the PALM semantic segmentation task to $1 \times 10^{-4}$ after 10 warum epochs. 
    \item \textbf{Weight decay}: in pretraining tasks, we use a weight decay factor of $1\times10^{-6}$. In downstream tasks, we use a weight decay factor of $1\times10^{-5}$.
    \item \textbf{Number of epochs}: in pretraining tasks, we train all models for 100 epochs. In downstream tasks, we fine-tune for:
        \begin{itemize}[noitemsep]
            \item For the PALM, APTOS, and RFMiD tasks: we train all models for 50 epochs.
            \item For Cardiovascular risk prediction tasks: we fine-tune all models for 5 epochs ($\approx 8000$ steps).
        \end{itemize}
    \item \textbf{Network architectures}: for the \textit{image encoder}, as mentioned before, we use a Resnet50~\cite{resnet} architecture across all pretraining and downstream tasks. For the \textit{genetics encoders}, we vary between following choices: 
        \begin{itemize}[noitemsep]
            \item None: here we do not have any hidden fully-connected layer for the genetics, and we feed them as inputs to the projection head directly.
            \item H1: we process the genetic inputs with one hidden layer of size 2048. (followed by a \texttt{ReLU} activation and \texttt{Batchnorm1D} layers)
            \item H12: we process the genetics with two hidden layers, both of size 2048. (Each layer is followed by a \texttt{ReLU} and \texttt{Batchnorm1D})
        \end{itemize}
    For the \textit{projection head}, we follow~\cite{simclr} in using two fully-connected layers. The first has a size of 2048 and is followed by a \texttt{ReLU}. The second has size of 128, which is the projection embedding size. Finally, for classification and regression downstream tasks we add one fully-connected \texttt{Linear} layer on top to perform the task. But for the \textit{PALM segmentation} task, we add a U-Net~\cite{UNET} decoder on top of the Resnet50 encoder. For upsampling layers in the decoder, we use transposed convolutional layers \texttt{ConvTranspose2d}. 
    \item \textbf{Loss functions}: the used loss functions for each task are as follows:
        \begin{itemize}[noitemsep]
            \item ContIG: for training our method, we use a contrastive loss (\texttt{NTXentLoss}). This loss is implemented using a cross-entropy loss, where the model is trained to classify which sample is positive in each mini-batch. However, our version of the \texttt{NTXentLoss} only does inter-modal contrasting, and not intra-modal. We set $\lambda=0.75$ in this loss (Eq. 1 in the main paper), and the temperature $\tau=0.1$. Note that a larger value of $\lambda$ gives more importance to image features than genetic features.
            \item APTOS \& RFMiD: we use the binary cross-entropy loss in both tasks. 
            \item PALM: we use a weighted combined loss of Dice-loss~\cite{dice_score} (weight=0.8) and binary cross-entropy (weight=0.2).
            \item Cardiovascular risk classification (sex \& smoking status): we use a binary cross-entropy loss.
            \item Cardiovascular risk prediction (age \& BMI \& SBP \& DBP): we use the Mean Square Error (MSE) loss. 
            \item SimCLR~\cite{simclr}: this method uses the contrastive \texttt{NTXentLoss} too. We similarly set the temperature $\tau=0.1$. 
            \item NNCLR~\cite{nnclr}: this method uses the contrastive \texttt{NTXentLoss} too. We similarly set the temperature $\tau=0.1$. 
            \item Simsiam~\cite{simsiam}: this method does not use negative sampling, and instead uses a Siamese network to minimize the similarity between two augmented views of the same image. Hence, the loss function used is the negative cosine similarity loss. 
            \item BYOL~\cite{byol}: this method has the same loss used in Simsiam, which is the negative cosine similarity. 
            \item Barlow Twins~\cite{barlow}: this method modifies the contrastive loss to compute the cross-correlation matrix between two sets of embeddings, which are for the same batch of images but with different image augmentations. Then, it tries to make this matrix close to the identity matrix. 
        \end{itemize}
\end{itemize}

\subsection{Implementation Details} 
We implement all of our methods using \texttt{Python}. The libraries we rely on are \texttt{PyTorch v1.9.1}, \texttt{Pytorch-Lightning v1.4.8}, \texttt{torchvision v0.10.0}, \texttt{torchmetrics v0.4.0}, and \texttt{Lightly}~\cite{lightly} (for baseline self-supervised implementations). We also follow the reproducibility instructions for \texttt{Pytorch-Lightning}~\cite{reproducibility}, \ie by setting a unified random seed of $42$ for all scripts and workers, and by using \texttt{deterministic} algorithms. We attach our source code with this supplementary material submission.

\section{Additional Downstream Results}

\subsection{Complete Finetuning Results}
In this section, we present the full set of results for fine-tuning our ContIG models versus the same baselines. These extended evaluation results, in \cref{tab:downstream_finetune}, show that ContIG is advantageous to the baselines. The rows in \cref{tab:downstream_finetune} are grouped in the following order: i) baseline trained from scratch, ii) self-supervised baselines, iii) ContIG trained on single genetic modalities with the images, and iv) ContIG trained on multiple genetic modalities with images. 

\subsection{Linear Evaluation Results} 
In this section, we follow a linear evaluation protocol~\cite{color,CPC1,simclr}, meaning that the encoder weights are kept frozen and we only train a linear classifier / regressor on top. As shown in \cref{tab:downstream_linear}, models trained with our method ``ContIG'' consistently outperform the baselines. Linear evaluation aims to provide a good idea about the quality of semantic representations stored in the model encoder.

\begin{table*} 
  \centering
  \begin{tabular}[t]{ l p{2cm} x{2cm} x{2cm} x{2cm} x{2cm} @{\hspace{-0.1cm}}c } \toprule
    \multicolumn{2}{c}{\multirow{2}{*}{Model \& Genetics Encoder}} & \multicolumn{1}{c}{APTOS} & \multicolumn{1}{c}{RFMiD} & \multicolumn{1}{c}{PALM} & \multicolumn{2}{c}{Cardio. Risk Pred.} \\
    \cline{3-7}
     & & QwKappa $\uparrow$ & ROC-AUC $\uparrow$ & Dice-Score $\uparrow$ & MSE $\downarrow$ & ROC-AUC $\uparrow$ \\
    \hline 
    Baseline                    & - & 80.47 & 91.64 & 77.25 & 3.440 & 56.29 \\
    \hline
    SimCLR~\cite{simclr}        & - & 81.83 & 91.88 & 70.41 & 3.451 & 59.38 \\
    SimSiam~\cite{simsiam}      & - & 75.44 & 91.28 & 72.26 & 3.442 & 57.37 \\
    BYOL~\cite{byol}            & - & 71.09 & 89.88 & 66.32 & 3.414 & 59.73 \\
    Barlow Twins~\cite{barlow}  & - & 72.28 & 92.03 & 70.53 & 3.430 & 59.05 \\
    NNCLR~\cite{nnclr}          & - & 77.93 & 91.89 & 72.06 & 3.426 & 61.95 \\
    \hline
    ContIG (Raw-SNP)            & None & 81.99 & 92.27 & 74.96 & 3.366 & 64.71 \\
    ContIG (Raw-SNP)            & H1   & 84.01 & 93.22 & 76.98 & 3.254 & 70.10 \\
    ContIG (Raw-SNP)            & H12  & 82.56 & 93.09 & 77.02 & 3.201 & 69.58 \\
    ContIG (PGS)                & None & 83.84 & 91.63 & 76.86 & 3.257 & 69.81 \\
    ContIG (PGS)                & H1   & \underline{85.93} & 93.31 & \textbf{78.47} & \underline{3.176} & \textbf{72.72}\\
    ContIG (PGS)                & H12  & \textbf{86.44} & 93.04 & 77.04 & 3.216 & 70.69 \\
    ContIG (Burden)             & None & 82.92 & \textbf{93.68} & 76.89 & 3.273 & 71.91 \\
    ContIG (Burden)             & H1   & 83.22 & 93.03 & 76.49 & \textbf{3.160} & \underline{72.37} \\
    ContIG (Burden)             & H12  & 83.61 & 93.14 & 76.72 & 3.236 & 71.50 \\
    \hline
    ContIG (Inner RPB)          & None & 83.49 & 93.31 & 77.11 & 3.195 & 71.68 \\
    ContIG (Inner RPB)          & H1   & 81.52 & 92.95 & 77.34 & 3.202 & 70.80 \\
    ContIG (Inner RPB)          & H12  & 80.24 & 92.94 & 75.37 & 3.235 & 68.89 \\
    ContIG (Outer RPB)          & None & 82.93 & 93.01 & 76.31 & 3.260 & 69.16 \\
    ContIG (Outer RPB)          & H1   & 84.22 & \underline{93.62} & 76.97 & 3.187 & 71.80\\
    ContIG (Outer RPB)          & H12  & 84.21 & 93.41 & \underline{77.51} & 3.233 & 71.13 \\
    \bottomrule
  \end{tabular}
  \caption{Downstream evaluation results by fine-tuning on each task. \textbf{Bold} indicates the best result, \underline{underlined} is second best. RPB in our method stand for the genetic modalities used: \textit{R}aw-SNPs, \textit{P}GS-scores, and \textit{B}urden-scores. $\uparrow$ means higher is better, and $\downarrow$ lower is better.}
  \label{tab:downstream_finetune}

  \centering
  \begin{tabular}[t]{ l p{2cm} x{2cm} x{2cm} x{2cm} x{2cm} @{\hspace{-0.1cm}}c } \toprule
    \multicolumn{2}{c}{\multirow{2}{*}{Model \& Genetics Encoder}} & \multicolumn{1}{c}{APTOS} & \multicolumn{1}{c}{RFMiD} & \multicolumn{1}{c}{PALM} & \multicolumn{2}{c}{Cardio. Risk Pred.} \\
    \cline{3-7}
     & & QwKappa $\uparrow$ & ROC-AUC $\uparrow$ & Dice-Score $\uparrow$ & MSE $\downarrow$ & ROC-AUC $\uparrow$ \\
    \hline 
    SimCLR~\cite{simclr}        & - & 35.02 & 86.53 & 59.77 & 3.998 & 52.26 \\
    SimSiam~\cite{simsiam}      & - & 21.25 & 87.91 & 56.58 & 3.998 & 53.13 \\
    BYOL~\cite{byol}            & - & 17.39 & 87.84 & 54.04 & 4.009 & 52.29 \\
    Barlow Twins~\cite{barlow}  & - & 44.75 & 87.65 & 59.52 & 3.952 & 54.28 \\
    NNCLR~\cite{nnclr}          & - & 24.76 & 85.80 & 66.25 & 3.870 & 54.17 \\
    \hline
    ContIG (Raw-SNP)  & None & 59.14 & 89.24 & 72.82 & 3.683 & 59.07 \\
    ContIG (Raw-SNP)  & H1   & 69.85 & 89.99 & 75.25 & 3.443 & 64.36 \\
    ContIG (Raw-SNP)  & H12  & 68.72 & 90.47 & 74.39 & 3.439 & 69.58 \\
    ContIG (PGS)      & None & 66.34 & 88.16 & 75.03 & 3.488 & 62.64 \\
    ContIG (PGS)      & H1   & \textbf{72.38} & 90.43 & 76.35 & 3.426 & 63.98 \\
    ContIG (PGS)      & H12  & 70.20 & 90.01 & \textbf{77.13} & 3.481 & 63.27 \\
    ContIG (Burden)   & None & 70.29 & \underline{91.08} & 75.31 & 3.453 & 64.72 \\
    ContIG (Burden)   & H1   & 70.67 & 90.62 & 75.42 & 3.421 & 64.70 \\
    ContIG (Burden)   & H12  & \underline{71.22} & \textbf{91.10} & 76.09 & 3.434 & \underline{64.84} \\
    \hline
    ContIG (Inner RPB)  & None & 70.26 & 89.94 & 75.27 & 3.439 & 63.84 \\
    ContIG (Inner RPB)  & H1   & 66.94 & 88.65 & 75.00 & 3.404 & 64.73 \\
    ContIG (Inner RPB)  & H12  & 68.41 & 90.56 & 73.08 & 3.457 & 63.45 \\
    ContIG (Outer RPB)  & None & 66.94 & 90.38 & 75.29 & 3.448 & \textbf{65.20} \\
    ContIG (Outer RPB)  & H1   & 66.60 & 89.46 & \underline{77.04} & \underline{3.398} & 64.59 \\
    ContIG (Outer RPB)  & H12  & 68.57 & 90.51 & 76.50 & \textbf{3.388} & \textbf{65.20} \\
    \bottomrule
  \end{tabular}
  \caption{Downstream evaluation results by linear evaluation on each task. Similarly, the results obtained by ContIG outperform all baselines. \textbf{Bold} indicates the best result, \underline{underlined} is second best. RPB in our method stand for the genetic modalities used: \textit{R}aw-SNPs, \textit{P}GS-scores, and \textit{B}urden-scores. $\uparrow$ means higher is better, and $\downarrow$ lower is better.}
  \label{tab:downstream_linear}
\end{table*}

\subsection{Data-Efficiency Results}
In this section, we assess the quality of semantic representations in a semi-supervised experimental scheme. We choose randomly 1\% and 10\% of the labels provided by UK Biobank (UKB)~\cite{ukbio}, and perform the downstream tasks of Cardiovascular Risk Factors prediction. Then, we evaluate using the same fixed test split of 20\% of UKB dataset size. We choose this particular downstream task as UKB's dataset size is large enough to allow a simulation for expert annotation collection process, \ie 1\% of number of overall labels is approximately 1000 samples, and such number may simulate an annotation process. The other benchmark datasets (APTOS~\cite{APTOS}, RFMiD~\cite{rfmid}, and PALM~\cite{palm}) are relatively small in size. 
The evaluation results shown in \cref{tab:data_efficient} compare models trained with ContIG to models trained with the self-supervised baselines. ContIG outperforms the baselines in this evaluation scheme too. Note that all models are trained on the same exact subset of individuals and also evaluated on the same test set. The results for this data-efficient evaluation scheme especially confirm the advantages of pretraining with multiple genetic modalities using the "Outer" aggregation scheme.
Notably, semi-supervised pretraining of ContIG with only 1\% labeled data still outperforms the self-supervised baselines when they have $10\times$ as much labeled data available.
\begin{table}[ht]
\small
  \centering
  \begin{tabular}[t]{ l c c c c } \toprule
    \multirow{3}{*}{Model} &  \multicolumn{4}{c}{Label Fraction} \\
     &  \multicolumn{2}{c}{1\%} &  \multicolumn{2}{c}{10\%} \\
    \cline{2-5}
     & MSE $\downarrow$ & ROC $\uparrow$ & MSE $\downarrow$ & ROC $\uparrow$\\
    \hline
    SimCLR~\cite{simclr}        & 4.029 & 51.43 & 3.762 & 54.29  \\
    SimSiam~\cite{simsiam}      & 3.861 & 53.35 & 3.564 & 57.45  \\
    BYOL~\cite{byol}            & 3.894 & 51.68 & 3.505 & 56.71  \\
    Barlow Twins~\cite{barlow}  & 3.788 & 51.89 & 3.558 & 56.86  \\
    NNCLR~\cite{nnclr}          & 3.913 & 52.20 & 3.643 & 55.99  \\
    \hline
    ContIG (Raw-SNP)            & 3.541 & \underline{60.11} & 3.414 & 64.81  \\
    ContIG (PGS)                & 3.521 & 59.23 & \underline{3.391} & \underline{65.86} \\
    ContIG (Burden)             & 3.540 & 59.74 & 3.393 & 65.41 \\
    ContIG (Inner RPB)          & \underline{3.511} & 59.95 & 3.397 & 65.71 \\
    ContIG (Outer RPB)          & \textbf{3.490} & \textbf{60.39} & \textbf{3.378} & \textbf{65.99}\\
    \bottomrule
  \end{tabular}
  \caption{Data-efficient evaluation results by fine-tuning on subsets of UKB samples. All our ContIG models use the "H1" genetic encoder variant. \textbf{Bold} indicates the best result, \underline{underlined} is second best.  $\uparrow$ means higher is better, and $\downarrow$ lower is better.}
  \label{tab:data_efficient}
\end{table}

\section{Additional Feature Explanation Results}
\subsection{Method Validation}
\begin{figure*}[ht]
  \centering
  \includegraphics[width=\linewidth]{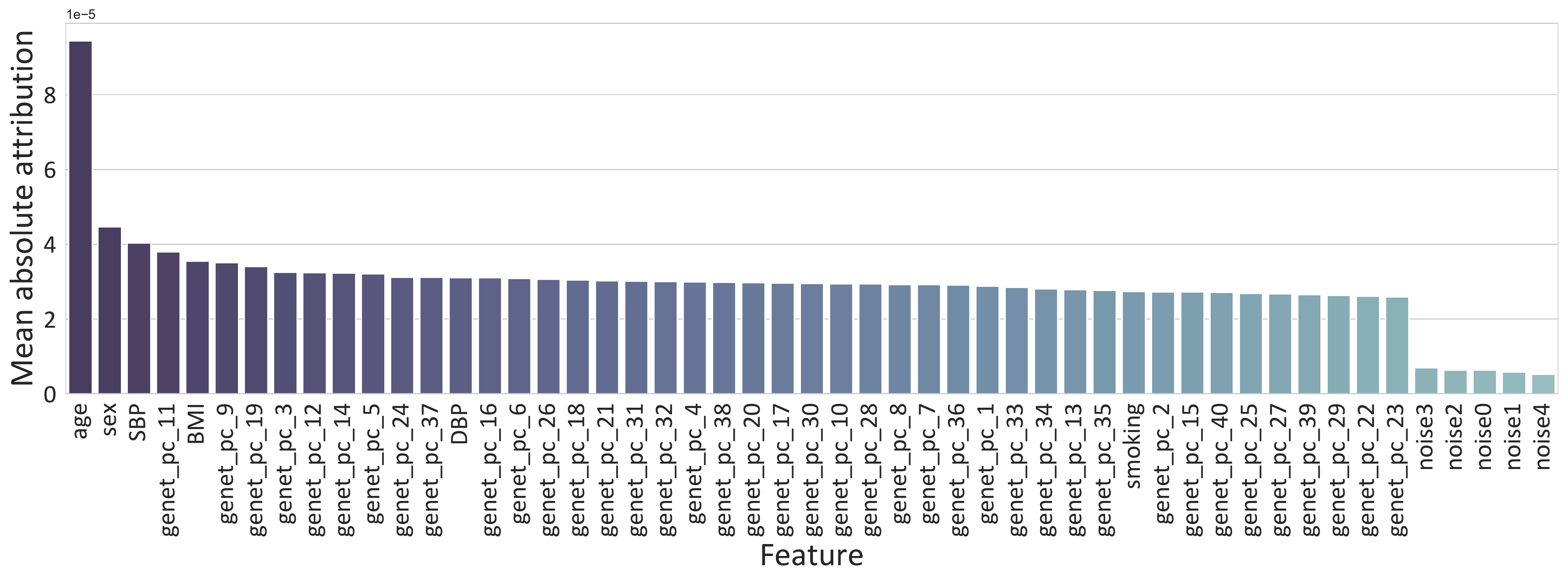}
  \caption{
  Explanation method validation. Shown is the mean absolute attribution for each feature aggregated over a batch-size of 1,000 individuals.
  \texttt{noise0, ..., noise4} don't carry any information and also get downweighted by our attribution method.
  }
  \label{fig:exp-validation}
\end{figure*}

We ran a baseline experiment to validate that our feature explanation method properly attributes to meaningful features.
In this experiment, instead of genetic features, we use phenotypic covariates such as age, sex, systolic and diastolic blood pressure (SBP and DBP), which can be predicted reliably from retinal fundus images.
Additionally, we include the first 40 principal components, which mostly capture population structure information.
As control variables, we also feed five random noise variables into the training process, which have no association with the images at all.
\cref{fig:exp-validation} shows the aggregated feature explanations.
As expected, the noise variables (\texttt{noise0, ... noise4}) get assigned very low explanation scores, while all other variables have considerable influence.
This validates that our feature explanation approach can distinguish between variables that carry true information relevant to the network and variables that are unrelated to the images.

\begin{figure*}[ht]
  \centering
  \begin{subfigure}[b]{0.49 \textwidth}
    \centering
    \includegraphics[width=\linewidth]{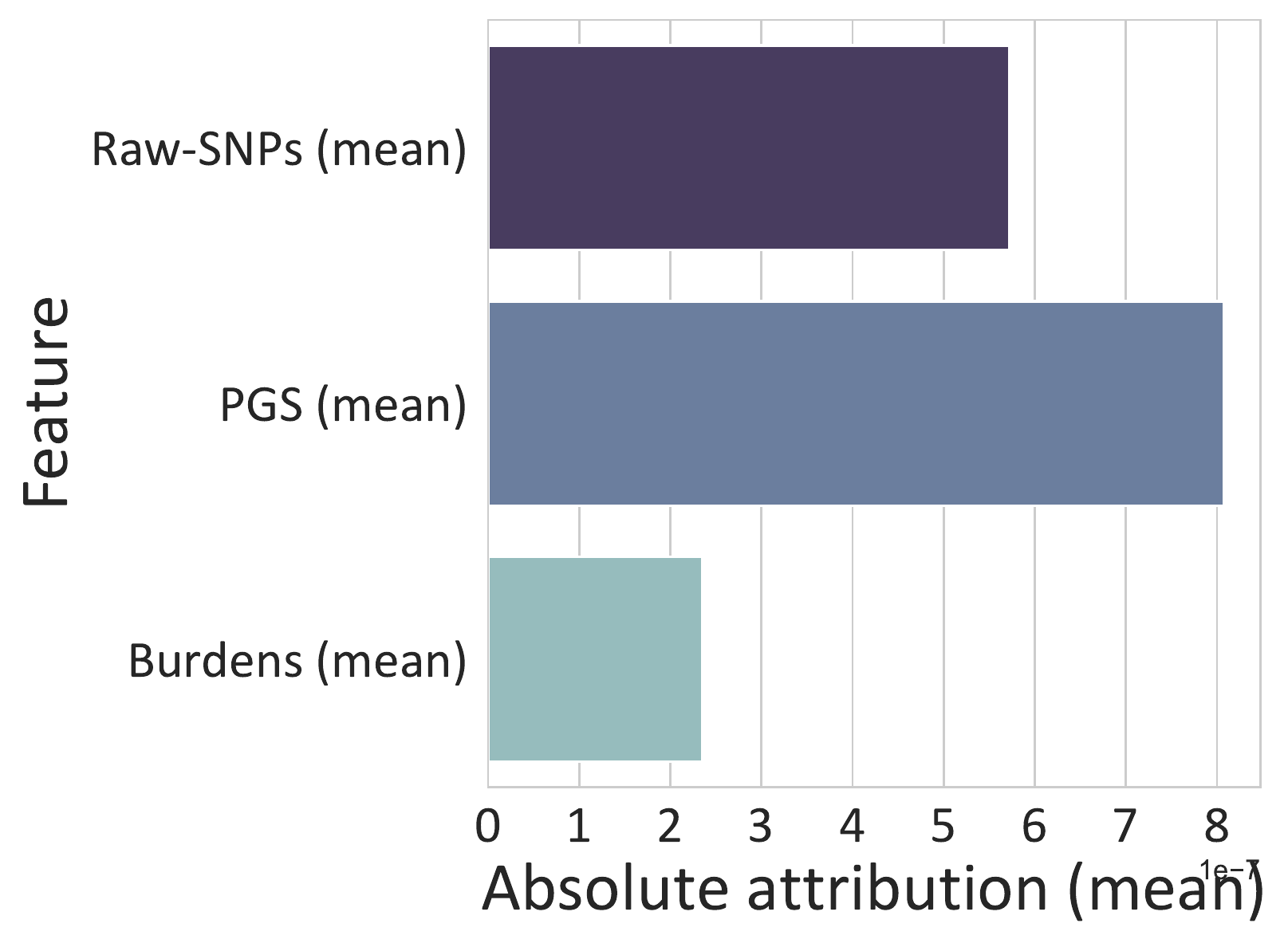}
    \caption{Absolute attribution for each modality, aggregated by mean.}
    \label{fig:exp-multi-mean}
  \end{subfigure}
  \begin{subfigure}[b]{0.49\textwidth}
    \centering
    \includegraphics[width=\linewidth]{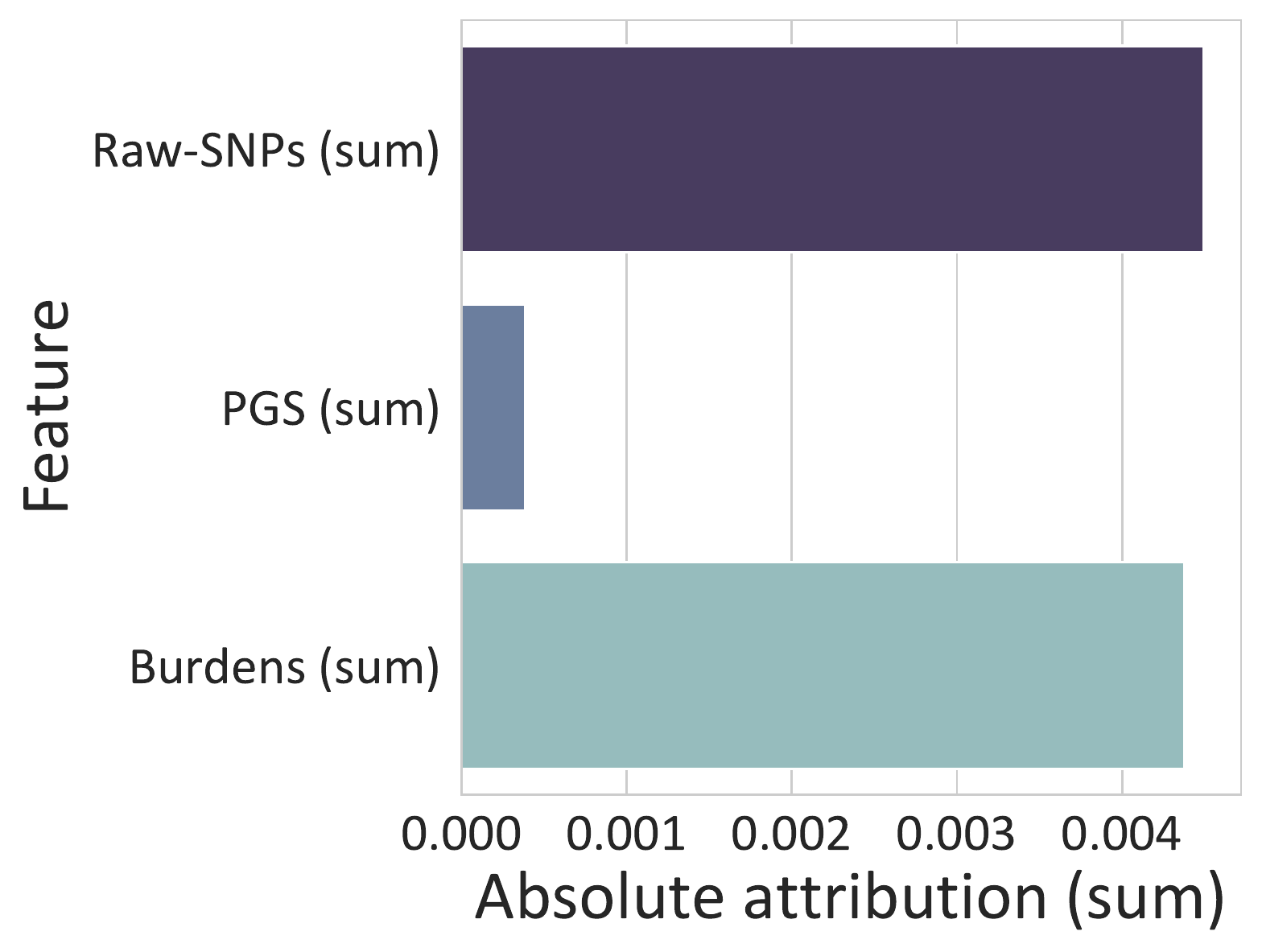}
    \caption{Absolute attribution for each modality, aggregated by sum.}
    \label{fig:exp-multi-sum}
  \end{subfigure}
  \caption{Absolute attributions by modality for ContIG (Outer RPB).}
  \label{fig:exp-multi}
\end{figure*}

\begin{figure*}[ht]
  \centering
  \includegraphics[width=\linewidth]{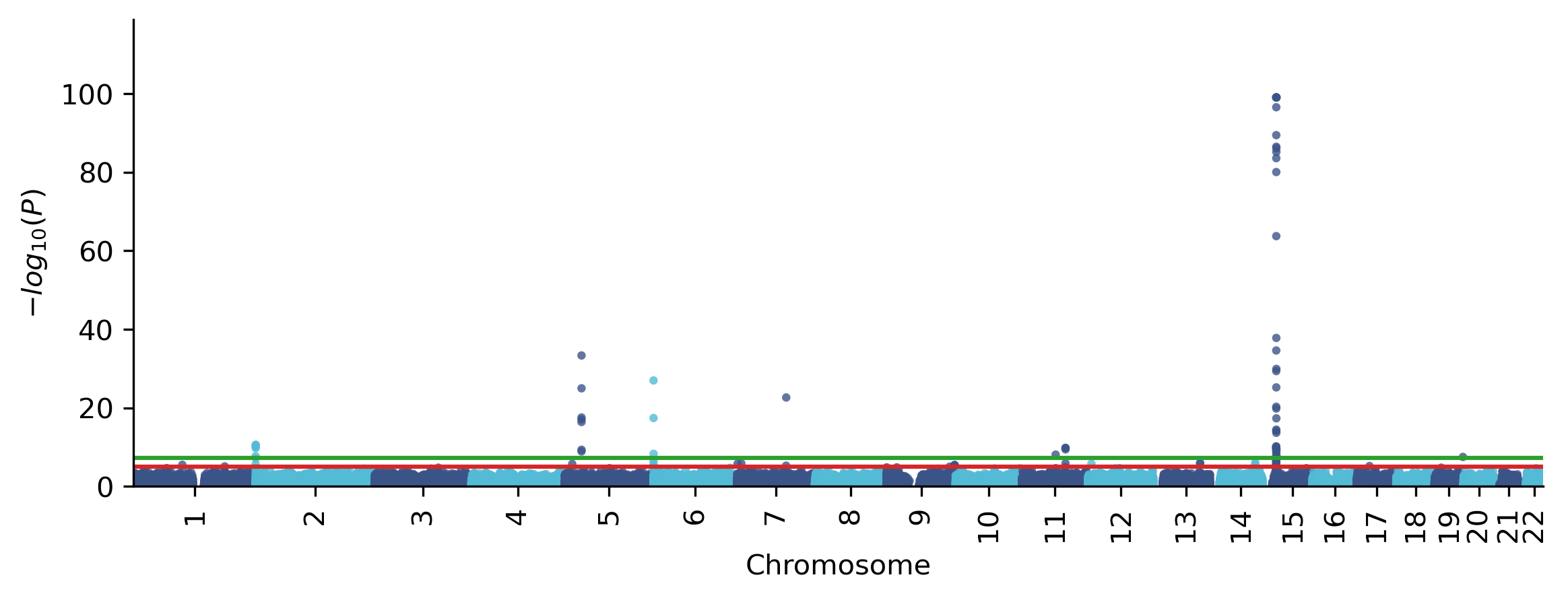}
  \caption{
    Manhattan plot for the GWAS with ContIG (Inner RPB).
    The x-axis shows the position of each SNP on the genome, the y-axis is the negative base-10 logarithm of the \textit{p}-value for each SNP.
    Higher values correspond to lower \textit{p}-values, correspond to stronger signal.
    The red line corresponds to a significance threshold of 0.05 Bonferroni-adjusted for the number of SNPs; the green line corresponds to ``genome-wide significance'' ($5 \cdot 10^{-8}$).
    \textit{P}-values are clamped at $10^{-99}$ for clearer visualization (only relevant for the loci on chromosome 15 with a minimum \textit{p}-value of $10^{-320}$).
    }
  \label{fig:mhat}
\end{figure*}

\subsection{Multimodal Explanation Results}
\cref{fig:exp-multi} shows the aggregated attribution scores for each of the three modalities, Raw-SNPs, PGS, and Burdens, for ContIG with the ``Outer'' training scheme.
\cref{fig:exp-multi-mean} shows that PGS scores on average have more influence than individual SNPs or burden scores.
However, \cref{fig:exp-multi-sum} also shows that that in aggregate, raw SNPs and burden scores have more total influence on the model.
This is likely due to PGS only having 481 features, while raw SNPs and Burdens have 7,854 and 18,574 features, respectively.
This may also explain the small but counterintuitive performance drop from ContIG (PGS) to ContIG (Outer RPB): the strongest signal, PGS, gets ``drowned out'' by the less important but overabundant signal in the raw SNPs and burden scores. 

\section{GWAS Analysis Details}

We produced feature vectors by computing the hidden-layer embedding for each image in the test-split of our dataset (10\% of the whole dataset, 7,079 individuals).
In contrast to the main training, we only used embeddings of the left eye and only included each individual once.
Feature vectors were reduced to 10 dimensions using a PCA.
Before computing the association results, we also used an inverse-normal transform \cite{sofer2019fully} after conditioning on the potential confounders ``sex'', ``age'', as well as the first 15 genetic PCs.
This ensures that the residuals of the marginal distributions are approximately normally distributed and outlier deviations from normality don't artificially inflate the type-1 error rate, leading to spurious correlations.
We performed the genetic association study with the PLINK2 software \cite{chang2015second}, using a linear model for each of the ten dimensions individually.
We again correct for the same confounders in the linear model.
Finally, we aggregate the summary statistics of the ten individual features into a single \textit{p}-value for each SNP by using a Bonferroni-correction of the factor 10, following \cite{transferGWAS}.

Genetic variants are locally highly correlated.
Therefore, we group significantly associated SNPs that are spatially close and in LD together using the PLINK \cite{purcell2007plink} clumping functionality (using parameters $\texttt{clump-p1} = 5\cdot 10^{-8}$, $\texttt{clump-p2} = 10^{-7}$, $\texttt{clump-r2} = 0.1$, $\texttt{clump-kb} = 150$).
We reported the number of independent associated regions returned by this procedure in the main document.

\cref{fig:mhat} shows the manhattan plot of genome-wide associations from the GWAS with ContIG (Inner RPB) pretraining.
A number of very strong signals, e.g. on chromosomes 15 and 5, are known to be associated with skin pigmentation and cardiovascular traits.
Manhattan plots for the other pretrained models look similar but with less signal.
Almost all models found the very strong signals on chromosome 15.
Interestingly, the manhattan plots for both SimCLR and BYOL (not displayed here) show clear signs of a ill-fitted association model, with many (for BYOL) but small, most likely spurious associations distributed over the whole genome but no signal in the chromosome-15 pigmentation region.
This happens even after applying the inverse-normal transformation to counteract outliers and is likely due to different forms of confounding.
This finding also explains the surprisingly large number of hits for BYOL -- they are most likely false-positives.
A more careful analysis with mixed effect models \cite{lippert2011fast} and in-depth inspection of the image features is beyond the scope of this article.

\end{appendices}

\end{document}